# Probabilistic Neural Network to Quantify Uncertainty of Wind Power Estimation

Farzad Karami, Nasser Kehtarnavaz, *Fellow, IEEE*, and Mario Rotea, *Fellow, IEEE*

*Abstract* — Each year a growing number of wind farms are being added to power grids to generate electricity. The power curve of a wind turbine, which exhibits the relationship between generated power and wind speed, plays a major role in assessing the performance of a wind farm. Neural networks have been used for power curve estimation. However, they do not produce a confidence measure for their output, unless computationally prohibitive Bayesian methods are used. In this paper, a probabilistic neural network with Monte Carlo dropout is considered to quantify the model (epistemic) uncertainty of the power curve estimation. This approach offers a minimal increase in computational complexity over deterministic approaches. Furthermore, by incorporating a probabilistic loss function, the noise or aleatoric uncertainty in the data is estimated. The developed network captures both model and noise uncertainty which is found to be useful tools in assessing performance. Also, the developed network is compared with existing ones across a public domain dataset showing superior performance in terms of prediction accuracy.

## I. INTRODUCTION

Wind is considered to be a major source of renewable energy around the world. The production of electricity from wind has been growing steadily in the last decade. In fact, the installed capacity of wind-produced electricity throughout the world has doubled from 2013 to 2018 [1].

Decisions required to install and operate wind plans are informed with the so-called "wind turbine power curve," which is a mathematical representation of power generated as a function of wind speed. This curve is provided by wind turbine manufacturers and is used to estimate the annual energy production (AEP). For example, the wind power curve can be used to assess the potential of developing a wind plant based on the wind speed profile of a given location [2]. Power curves can also be used by the operator to identify a faulty behavior or abnormal operation of a turbine by comparing the actual generated power with the power curve [3].

In practice, however, the actual power generated by a wind turbine deviates from its nominal power curve as there are variations and uncertainties associated with wind velocity (speed and direction) under real-world operating conditions. Hence, given the role played by power curves in the development and operation of wind plants, its continuous accurate estimation is of practical importance.

Several methods have been proposed for estimating a power curve. These methods can be categorized into parametric and non-parametric models. Parametric models involve a few well-defined parameters. Piecewise linear regression, nonlinear regression, and piecewise splines are examples of parametric models applied to the power curve estimation problem [4, 5]. Examples of nonparametric models include support vector machine (SVM), k-nearest neighbor (KNN), random forest (RF), gradient boosting (GB), and neural network (NN) [2, 6, 7]. The power curve estimation methods can also be grouped into deterministic and stochastic ones. Deterministic methods involve point prediction of the output while stochastic methods seek a probability distribution of the output [2, 8]. Due to the random nature of atmospheric variables, as well as various factors affecting a wind turbine, there is uncertainty associated with the power curve estimation [9]. It is therefore beneficial to quantify this uncertainty which can then be utilized in the operation of wind turbines. The estimated uncertainty can help planners and wind farm operation engineers to have a better view of the validity of the wind power curves, hence enhancing their confidence to make decisions. Stochastic methods attempt to provide a probability distribution associated with the output power. The Gaussian Process Regression (GPR) method has been previously used to acquire uncertainty in the power curve [10-14]. Resampling methods such as bootstrapping and bagging have been used to obtain uncertainty in terms of confidence intervals [15]. Uncertainty propagation methods have been considered to construct a probability density function based on the uncertainty of input variables via curve fitting [16].

A shortcoming of deterministic methods is that they do not take into consideration the distribution associated with the input variables and the output is found as a single point or value. In other words, one does not know the uncertainty associated with an output given an input. Although stochastic methods attempt to address this issue, they have limitations. The GPR method performs well for small to medium size datasets, its performance degrades for large datasets due to the computation of the covariance kernel that is needed for all the dataset samples. Among the aforementioned methods, neural networks are being increasingly used for power curve estimation due to their ability to estimate the nonlinearities involved [7, 17]. In [18], it is shown

---

This research was partially supported by NSF Award No. 1839733 from the CMMI division of ENG directorate.
Farzad Karami is with Mechanical Engineering Department, University of Texas at Dallas, Richardson, TX 75080 USA.
Nasser Kehtarnavaz is with Electrical and Computer Engineering Department, University of Texas at Dallas, Richardson, TX 75080 USA.
Mario Rotea is with Mechanical Engineering Department, University of Texas at Dallas, Richardson, TX 75080 USA.

that neural networks perform better than conventional methods, such as polynomial regression, logistic regression, and KNN, for the power curve estimation problem. Neural networks, in their basic form, are deterministic models. To take into consideration the uncertainty associated with the estimated output power, the Bayesian approach can be deployed. However, the use of the Bayesian approach is computationally intensive and time-consuming [19].

This paper addresses the uncertainty associated with the power curve estimation. For this purpose, a probabilistic neural network model based on the Bayesian approach is developed in order to obtain the uncertainty associated with the inevitable modeling error, which is called *epistemic uncertainty*, in a computationally efficient manner. The high computational demand of the Bayesian approach is mitigated here by using the method of Monte Carlo dropout. Furthermore, by modifying the loss function of the network, the *aleatoric* uncertainty associated with data noise is estimated. The calculated uncertainty helps developers in two ways. The epistemic uncertainty shows the quality of the model development process including the goodness of the training dataset and sufficiency in the complexity of the implemented model. This quantity can be used to improve the model training process. On the other hand, the aleatoric uncertainty helps quantify the noise level in the data.

In this paper, the proposed model is trained and tested on a dataset from an existing wind farm. Both the accuracy and the uncertainty aspects of the output power estimation are examined by carrying out a comparison with existing methods. In the first section of this work, a brief introduction to the topic is provided along with the research question addressed in this paper. Then, the method, dataset, and training process are explained. This description is followed by the results section. In the final section, the findings are summarized and a path for further research is proposed.

II. DEVELOPED PROBABILISTIC NEURAL NETWORK MODEL

This section discusses the mathematical aspects of the developed neural network model, and the dataset used for training and testing the model.

*A. Mathematical Framework*

This section covers our methodology to estimate the probability distribution of wind power curves using measurements from a typical wind plant SCADA system. Two sources of uncertainty are identified in statistical learning problems. The first one, called epistemic uncertainty, arises from the lack of knowledge in the model regarding the actual nature of the model subject. Lack of enough data in certain areas of the feature space is the root cause of this uncertainty in machine learning models. The model uncertainty can be mitigated by providing more data for training the model to cover all the possible behaviors of the model. Of course, it comes at a price that is longer training time and more complex models to learn those data. The next source of uncertainty is the presence of noise in measurements. This type of uncertainty, called aleatoric uncertainty, can be reduced by using better measurement sensors. It can partly be done by appropriate postprocessing (e.g., de-noising) of the dataset. Note that modification of sensors on a complex system, such as wind turbines, can be a prohibitive task. A comprehensive probabilistic model is considered here to quantify these uncertainties [20, 21]. Neural networks are shown to provide an effective estimation model. Due to their flexible structure, they can estimate any nonlinear function by having adequate numbers of neurons and layers. They can handle a large number of samples in training datasets as well as high-dimensional datasets [17]. However, implementing a probabilistic approach is not trivial for a neural network model. The reason is the large number of parameters involved and also the highly nonlinear structure of these models, which make both analytical and numerical manipulation a demanding task for finding posterior distributions.

Recently, there have been several proposals for simplifying the development of probabilistic neural networks. Among them, the Monte Carlo Dropout method is applied to several applications due to its accuracy in determining output distributions and also convenient implementation. In its original form, this method is an approximation for a Bayesian posterior of the parameters in a neural network model. A summary of this method is given next.

A probabilistic model aims to estimate the posterior predictive function which is defined as follows [17, 22]:

$$p(y_{new}|D) = \int p(y_{new}|D,W) p(W|D) dW \quad (1)$$

where $y_{new}$ denotes the model prediction, $D = (X, Y)$ is the training dataset, and $W$ represents the model parameters. To obtain the above integral, one should have the posterior distribution of the parameters given the training data. By using the Bayes theorem, one could rewrite this distribution as follows [22]:

$$p(W|D) = \frac{p(D|W) p(W)}{p(D)} \quad (2)$$

Given the large number of parameters in a multilayer neural network model and the nonlinear relationship due to the existence of activation function, computing the above equations is demanding. In this method, the posterior pdf of the model parameters is approximated by variational inference through minimizing the Kullback–Leibler (KL) divergence between an approximate distribution $q_\theta(W)$, parametrized by a latent variable $\theta$, and the actual posterior distribution of $p(W|D)$ [23, 24]. Using the definition of KL divergence, one can minimize the following function with respect to $\theta$ to get $q_\theta(W)$ as an approximation of $p(W|X)$,

$$L(\theta) = -\int q_\theta(W) \log p(Y|X,W) dW + KL(q_\theta(W) \| p(W)) \quad (3)$$

The integral in (3) is still intractable and as an approximate solution, the Monte-Carlo integration can be adopted. At each step of minimization, the network output is evaluated using a sample drawn from $q_\theta(W)$. Repeating this process in the training phase optimizes the loss function.

The key concept for establishing $q_\theta$ is to convert the variational variable in the $q_\theta$ pdf into a random variable. For this purpose, dropout layers are considered. Dropout is a method of setting a certain number of network weights to zero at each training epoch via sampling from a Bernoulli distribution [25].

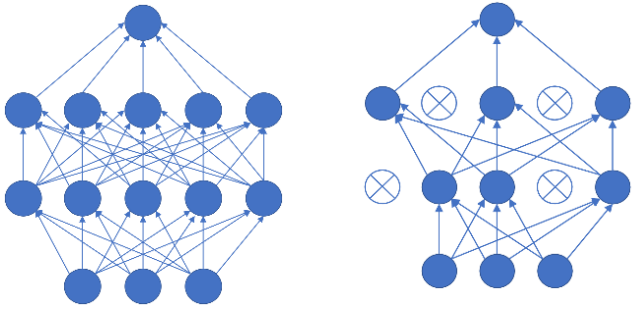

FIGURE 1. A feedforward neural network (left) versus its dropout processing unit version indicated by crosses (right) with a 0.4 dropout probability.

It is assumed that $q_\theta$ can be factorized over layers, that is:

$$q_\theta(W) = \prod_i q_{\theta_i}(W_i) \quad (4)$$

where $i$ denotes the layer number. For each layer, $q_\theta$ is parametrized to be the mean value of the weight matrix multiplied by a diagonal matrix consisting of 0 and 1 sampled from a Bernoulli distribution with a probability of $p_i$. Hence, some of the weights are randomly set to zero. As a result, the argmax of equation (3) becomes the mean value of the weights minimizing the loss function. In [26], it has been shown that the KL divergence can be estimated via the following equation:

$$KL(q(W_i) \| p(W_i)) \propto \frac{l^2 p_i}{2} \|W_i\|^2 - kH(1 - p_i) \quad (5)$$

where $l$ denotes the characteristic length (a hyperparameter), $k$ is the number of processing elements at $i^{th}$ layer, $H$ is the entropy of the Bernoulli distribution with probability $p_i$, and the subscript $i$ denotes the layer number. Therefore, a loss function for the network is established which consists of the log-likelihood of the training samples and the regulator term of (5) to keep $q_\theta$ close to the true posterior pdf of $p(W|D)$. As the gradient of (3) is needed for minimization, the problem with the discontinuity of the Bernoulli distribution, appearing in the entropy term of (5), needs to be addressed. For this purpose, a continuous approximation of the Bernoulli distribution is adopted [27]. This way, by minimizing the loss function of the network, the best probability of the dropout Bernoulli distribution is found [26].

With an assumption of the Gaussian distribution of the noise, the aleatoric uncertainty can be estimated through the training process via modifying the loss function [28, 29]. The logarithm of this term is substituted in (3) in place of log $p(Y|X,W)$,

$$L = \sum_{i=1}^{N} \frac{1}{2} \left( \frac{[y_i - \hat{y}_i]^2}{\sigma_a^2(x_i)} + \ln\left[\sigma_a^2(x_i)\right] \right) \quad (6)$$

where the subscript $a$ indicates aleatoric uncertainty. We use the loss function of (6) as the kernel for integrating (3).

The predictive probability of (1) can be approximated by a sampling of $W$ and feedforward predictions of the network. The mean value of this feedforward evaluation is computed as follows:

$$\bar{y} = \frac{1}{B} \sum_{i=1}^{B} \hat{y}_i(x_{new}, W) \quad (7)$$

where $B$ denotes the number of feedforward evaluations and $\hat{y}_i$ the output of the model at $i^{th}$ evaluation. The corresponding variance is computed as stated below.

$$\sigma_e^2 \simeq \frac{1}{B} \sum_{i=1}^{B} \left(\hat{y}_i(x_{new}, W_i^s) - \bar{y}\right)^2 \quad (8)$$

The above variance is an indicator of the epistemic uncertainty, denoted by subscript $e$, and it arises from the uncertainty in the model parameters updated by a training dataset, i.e. $p(W|D)$.

B. Network Architecture

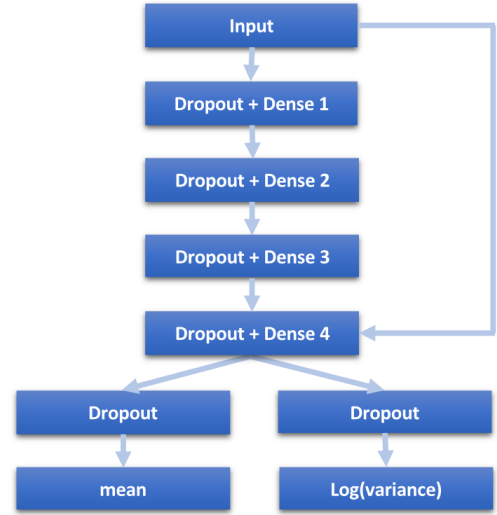

FIGURE 2. Architecture of the probabilistic neural network model used

A multi-layer neural network is used here by stacking four fully connected layers each consisting of 1024 processing units and two parallel layers with 1 processing element for the mean and aleatoric variance outputs at the network end. The input is also feedforwarded to the fourth layer. The architecture of the network is illustrated in FIGURE 2. The activation function $tanh(\cdot)$ is used for the first four layers and linear activation functions are implemented for the last layers. Dropouts are placed between the layers with the Bernoulli distribution probability set by the optimization algorithm. The mean absolute error (MAE) is used for evaluating the performance of the model. This metric indicates how a model performs over a dataset and is defined as follows: [22]

$$MAE(S) = \frac{1}{N}\sum_{i=1}^{N}|y_i - \hat{y}_i| \qquad (9)$$

where $N$ is the number of dataset samples, $S$ is the test dataset, $y_i$ is the true output of the $i^{th}$ sample and $\hat{y}_i$ is the model output for the same sample.

The Adam optimizer algorithm is chosen for minimizing the loss function in equation (3) with a learning rate scheduling plan starting from 1E-3 and ending at 1E-5. After the training phase, the estimation is done for the test set having no overlap with the training set. For evaluating the epistemic uncertainty distribution, the feedforward output evaluation is repeated for each input sample and the results are recorded. Then, the mean value of the distribution is taken as the estimated power value and the standard deviation is considered to be the epistemic uncertainty. The logarithm of the variance, computed by the log(variance) layer of FIGURE 2, is a measure of aleatoric uncertainty. The code is written in Python using the utility Keras with TensorFlow 2.0 backend. The PC used for running the code has a 64 GB of RAM, a i7-7700K CPU operating at 4200 GHz, and a Quadro P4000 GPU.

*C. Datasets*

In this section, the datasets used are described. These datasets are from the wind turbines in La Haute, France [2], which are publicly available and consist of 5 years of SCADA data of 4 wind turbines. The data are aggregated for every 10 minutes and the mean, standard deviation, minimum, and maximum values are provided as time series. The turbines are MM82 and their specifications are shown in TABLE I. The nominal power curve of the turbine furnished by the manufacturer is used as the baseline to examine performance.

TABLE I
Specifications of MM82 wind turbine [30]

| Datasheet SENVION MM82 | |
| --- | --- |
| Nominal Power | 2050 [kW] |
| Rotor Diameter | 82 [m] |
| Blades Length | 40 [m] |
| Cut-in Wind Speed | 3.5 [m/s] |
| Cut-out Wind Speed | 20 [m/s] |
| Rated Wind Speed | 14.5 [m/s] |

A list of measurements and their description can be found on the website mentioned on the host website [31]. The wind speed was directly measured by two anemometers on the tower at different heights. As illustrated in FIGURE 3 (top), the joint distribution of the generated power and the wind speed is shown for a sample time interval. The two marginal distributions are shown on the top and right sides of the scatterplot for the wind speed and power, respectively. As indicated in this figure, the wind speed average is 6 m/s, and the power average lies around 450 KW. FIGURE 3 (bottom) shows a wind rose illustration for the same period, indicating that the southwest wind is dominant followed by the northeast wind with relatively very low winds in the other directions.

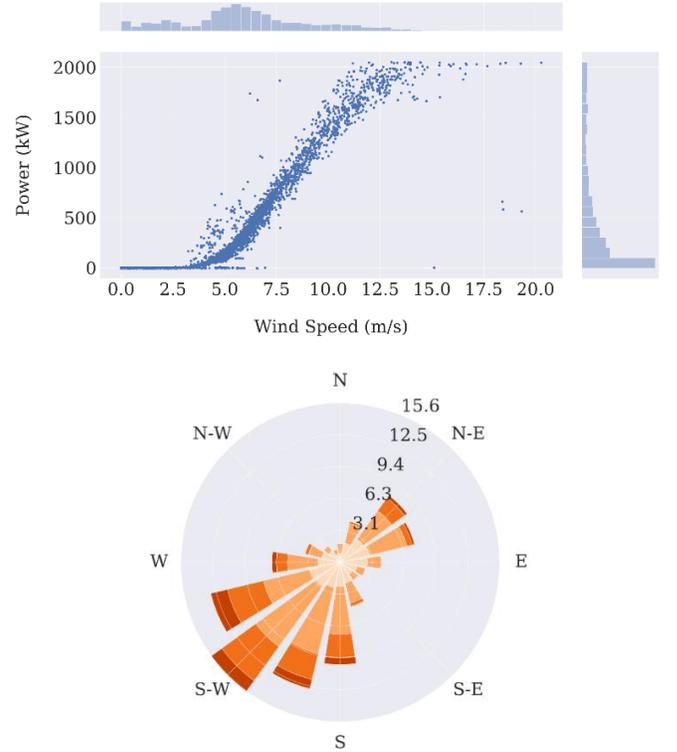

FIGURE 3. Joint distribution of wind speed and wind power for a typical wind turbine (top), Wind rose illustration at the wind turbine anemometer (bottom).

Wind speed is the primary predictor of wind power. However, other atmospheric variables such as air temperature have been found to contribute to the generated power [32]. The wind direction plays a role in the generated power, particularly if there is yaw misalignment. Here, the direction angle is transformed by sinusoidal functions and used as independent variables. The nacelle angle of the turbine is also reported in the datasets. The pitch angle of the blades affects the power output as well. These parameters are included in the wind power prediction. The wind turbulence intensity has an impact on the amount of energy a turbine can harvest. An increase in the turbulence intensity causes the generated power to deviate from the nominal curve. Thus, for accurate power estimation, it is important to consider this quantity [33, 34]. Turbulence intensity is defined as follows:

$$\tau = \frac{\sigma_v}{\bar{v}} \qquad (10)$$

where $\bar{v}$ denotes the average wind velocity and $\sigma_v$ the standard deviation of the wind speed on a given time interval. The other quantity affecting the output power is wind shear, which is defined as the ratio of wind speed at different heights [34]. The definition of wind shear is given by:

$$\alpha = \ln\left(\frac{v}{v_0}\right)\left(\ln\left(\frac{z}{z_0}\right)\right)^{-1} \qquad (11)$$

where $v$ denotes the velocity at height $z$ and $v_0$ the velocity at height $z_0$. Due to the unknown height of the anemometers, the ratio of $v$ to $v_0$ is taken here to serve as the indicator of the wind shear. The wind gust factor $G$ is another quantity that may impact power generation. It is an indicator of how large the magnitude of the wind bursts are compared to the average wind speed in a given time interval [34],

$$G = \frac{v_{\max}}{\bar{v}} \qquad (12)$$

The following values are used to form the input of the neural network: 10 minutes average wind speed, ambient temperature, wind direction, blades pitch angle, nacelle angle, wind turbulence intensity, wind gust, and wind shear. The effect of each one is investigated in the results and discussion section. The statistics of these inputs are listed in TABLE II.

TABLE II.
Typical input values to the neural network model.

| Variable | Mean | STD |
|---|---|---|
| Wind Speed (m/s) | 6.449 | 2.909 |
| Temperature (°C) | 4.953 | 3.456 |
| Wind Direction (deg) | 173.93 | 87.93 |
| Turbulence Int. | 0.133 | 0.096 |
| Gust Factor | 1.381 | 0.0317 |
| Wind Speed Ratio | 1.042 | 0.069 |
| Pitch Angle (deg) | 9.72 | 24.51 |
| Nacelle Angle (deg) | 174.46 | 88.44 |

### III. RESULTS AND DISCUSSION

In this section, the results of the developed model for estimating the wind power curve are presented. The model is evaluated in two ways. First, its effectiveness in evaluating uncertainty is studied. Noting that the main predictor of wind power is wind speed, for better visualization, the discussion focuses on the wind speed distribution rather than the overall joint input set distribution.

In FIGURE 4 the epistemic uncertainty of the test dataset is depicted. Each dot in the figure shows the mean wind power computed according to (7) and its hue indicates the corresponding variance obtained from (8). As illustrated in this figure, the uncertainty is the lowest in the region with wind speeds between 5 m/s and 8 m/s. This region corresponds to the highest probability region shown in FIGURE 3. The uncertainty of the model increases at the two extremes ends of the wind speed, particularly at higher speeds. This implies that the model estimation capability is capped as a result of fewer data points in those regions. Also, the data points, which are farther than the main body of the dataset, exhibit larger epistemic uncertainty.

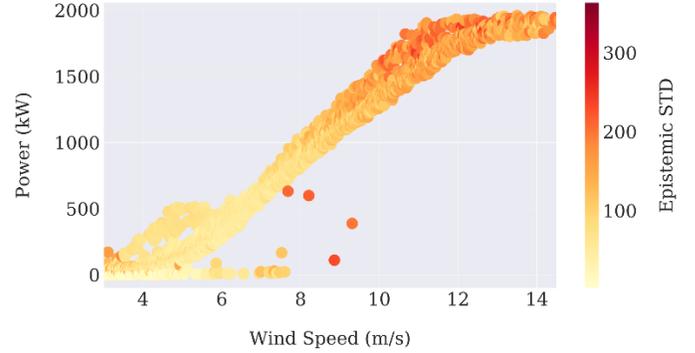

FIGURE 4. Model uncertainty for the Mont Carlo dropout neural network. Darker hue shows higher uncertainty as the data in the training dataset is more scarce compared to the lighter region corresponding to areas having more training data.

Next, the variation of the epistemic uncertainty with the relative frequency of the wind speed distribution in the training set and then the total number of samples used for training. In FIGURE 5 the samples are binned into fixed intervals of wind speed and then the average epistemic uncertainty for each bin is plotted against the frequency of the same bin. This figure shows that the epistemic uncertainty decreases with the frequency of the samples in the input dataset.

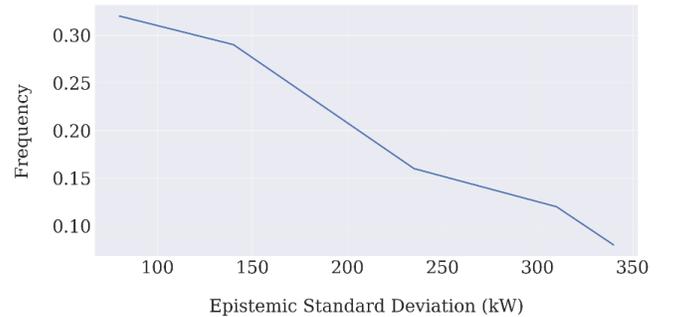

FIGURE 5. Epistemic standard deviation of generated power vs. frequency of wind speed magnitude in the dataset examined

The next study shows the performance of the model in evaluating the aleatoric uncertainty. This is the direct output of the model from the definition of the loss function. FIGURE 6 illustrates this quantity for the test dataset. The computed aleatoric uncertainty is seen higher for the samples with lower joint input-output probability. The algorithm treats these samples as noise and adjusts the standard deviation in the loss function to a higher value to reduce their effect on the output. Almost all the high uncertainty points are from data points far from the main body of the power curve. The opposite in the epistemic uncertainty plot is observed as the algorithm assigns low uncertainty to the points with wind speed between 4 m/s and 6 m/s and power of around 500 kW where they have the largest aleatoric uncertainty.

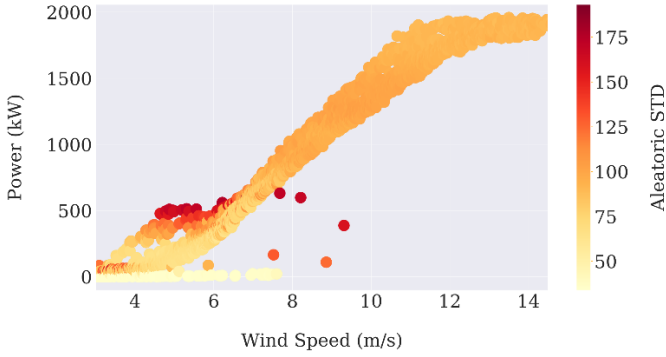

FIGURE 6. Aleatoric uncertainty for the test dataset. The standard deviation is higher for the samples located further from the concentration region.

The rest of this section is dedicated to examining the accuracy of the developed model in determining wind power, regardless of its ability to determine the uncertainty. First, the effect of each input on the accuracy of the power estimation is examined. The wind speed is taken as the main predictor of power and the percentage improvement in the accuracy in terms of MAE of the developed model is reported when adding inputs. In FIGURE 7 the mean absolute error for the model trained with different input variables is shown. For each case, the model was trained on three separate datasets and the results are averaged for the plot shown in this figure.

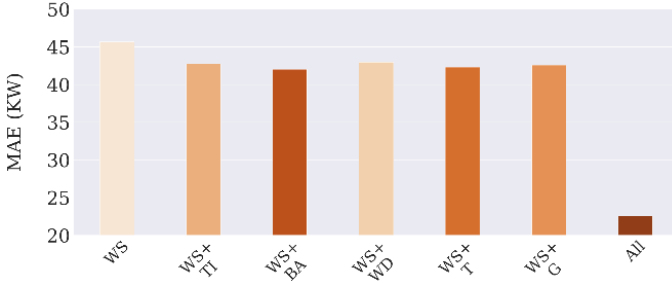

FIGURE 7. Mean absolute error (MAE) of the developed Monte Carlo dropout neural network model for different sets of inputs; the highest accuracy is achieved when all the inputs are considered. (WS: wind speed, TI: turbulence Intensity, BA: blade angle, T: temperature, G: gust)

TABLE III provides the performance comparison of different models on four datasets in terms of mean absolute error (MAE) in power estimation. The Gradient Boosting (GB), Random Forest (RF), and Gaussian Process Regression (GPR) models were examined in addition to a deterministic Neural Network (NN) for comparison with our developed MC dropout NN model. The results show that the developed model outperforms the other models across all the datasets.

TABLE III.
Comparison of the mean absolute error of the developed model with the other models consisting of gradient boosting (GB), random forest (RF), Gaussian process regression (GPR), deterministic deep learning (Vanilla NN). The estimations for the RF and GB methods are found through a grid search for the best mean absolute error on the validation dataset.

| Dataset Number | GB | RF | GPR | Vanilla NN | MC Dropout NN |
|---|---|---|---|---|---|
| 1 | 24.01 | 24.09 | 32.03 | 32.61 | **22.83** |
| 2 | 24.82 | 23.49 | 32.01 | 24.86 | **21.11** |
| 3 | 17.48 | 21.03 | 27.12 | 26.00 | **17.23** |
| 4 | 25.03 | 28.04 | 35.97 | 33.17 | **25.00** |

Dataset 1: 2017 Summer
Dataset 2: 2016 Winter
Dataset 3: 2014 All Seasons
Dataset 4: 2014 and 2015 All Seasons

FIGURE 8 shows the distribution of generated power predicted by the developed model along with the distribution of the actual power from the test dataset for the same inputs. Also, the power predicted by the manufacturer curve for the corresponding wind bin center is shown by black solid lines. Note that the developed model successfully constructs the distribution of the power for the unseen input data.

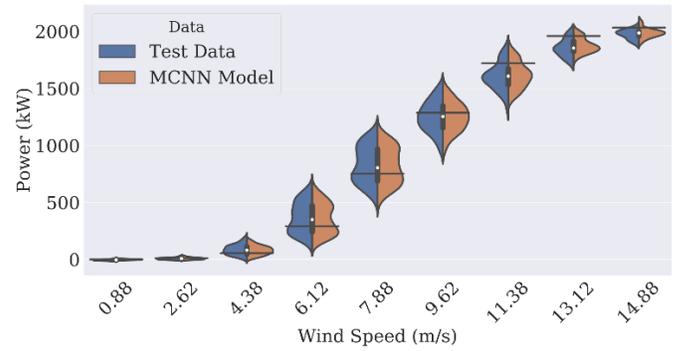

FIGURE 8. Distribution of the generated power from the test dataset and calculated by the model shows a good prediction power.

Finally, the percentage of improvement compared to power determination using only the nominal power curve is reported. In FIGURE 9, the nominal curve provided by the company and actual data are displayed on the top and the percentage improvement in mean absolute error on a dataset comparing to the nominal curve are displayed on the bottom.

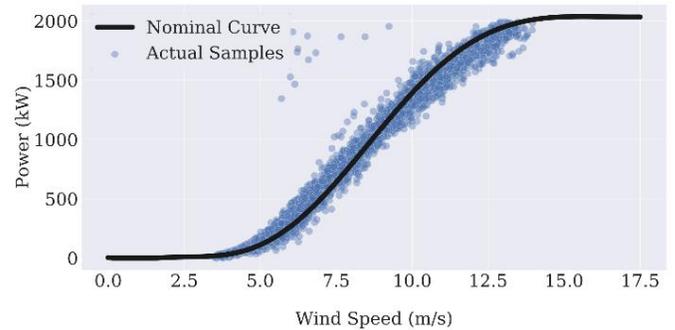

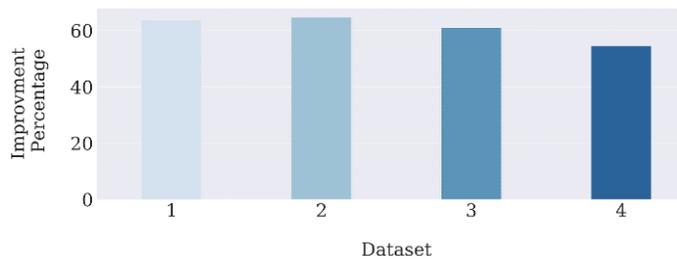

FIGURE 9. Turbine nominal curve and actual data (left), and percentage improvement in the mean absolute error of power estimation with MC dropout neural network model compared to the nominal curve (right) for each dataset.

## IV. CONCLUSION

In this paper, a method for determining the uncertainty of wind power estimation has been presented. Both epistemic and aleatoric uncertainties are obtained. For the epistemic uncertainty, the Monte Carlo dropout method is deployed as part of a regression neural network model to obtain the uncertainty of the model parameters. Then, by modifying the loss function of the network, the aleatoric uncertainty is found in the form of the output variance. The developed method combines the strength of neural networks with the advantages of Bayesian probability inference without suffering from the prohibitive computational complexity of the latter. Due to the dropout layers, this model is mostly immune to overfitting. The performance of the model was tested on four datasets from an actual wind farm. The epistemic uncertainty was found to be mostly due to the lack of enough information in the training datasets as the model showed less confidence in its output for samples from sparse input space regions. On the other hand, the aleatoric uncertainty, which originates from the noise in data samples, was seen to be able to identify outliers by assigning higher uncertainty to those samples. The model performance was compared to four other machine learning algorithms and the results showed the higher accuracy of the Monte Carlo dropout method in terms of the mean absolute error in the output power.

As future works, there is room for further exploitation of Bayesian neural networks. For instance, using the proposed method for the prediction problem of an entire wind plant can provide confidence intervals for the predicted power towards more reliable planning as it provides a tool for assessing annual power prediction with associated uncertainty. Also, the ability of this method in identifying outliers can be used for anomaly detection, fault detection, and condition monitoring of wind power systems.

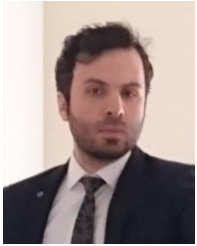

**Farzad Karami** received the BS and MS degrees in mechanical engineering from the Sharif University of Technology, Tehran, Iran, in 2009 and 2012 respectively. He is currently working toward a Ph.D. degree at the University of Texas at Dallas. His research interest includes the mathematical and statistical modeling of mechanical, electrical, and physical systems.
He had been working as an R&D and design engineer for 6 years before joining the University of Texas at Dallas. He has published over 12 journal and conference papers.

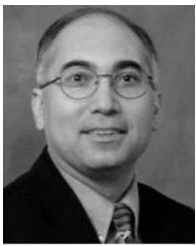

**Nasser Kehtarnavaz** (S'82–M'86–SM'92– F'12) is an Erik Jonsson Distinguished Professor with the Department of Electrical and Computer Engineering and the Director of the Embedded Machine Learning Laboratory at the University of Texas at Dallas, Richardson, TX. His research interests include signal and image processing, machine learning, deep learning, and real-time implementation on embedded processors. He has authored or co-authored ten books and over 400 journal papers, conference papers, patents, manuals, and editorials in these areas. He is a Fellow of SPIE, a licensed Professional Engineer, and Editor-in-Chief of the Journal of Real-Time Image Processing.

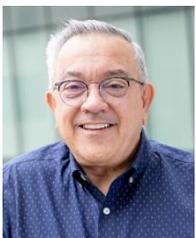

**Mario A. Rotea** (F'07) is the holder of the Erik Jonsson Chair in Engineering and Computer Science at the University of Texas at Dallas, where he is also the department head of mechanical engineering. Rotea spent 17 years at Purdue University as a professor of aeronautics and astronautics, developing and teaching methods for the analysis and design of control systems. He also worked for the United Technologies Research Center as a senior research engineer on advanced control systems for helicopters, gas turbines, and machine tools. Rotea was the head of the Mechanical and Industrial Engineering Department at the University of Massachusetts Amherst, where he expanded the department in the area of wind energy and applications of industrial engineering to the health care sector. His career includes terms as director of the Control Systems Program and division director of Engineering Education and Centers at the National Science Foundation. Rotea is cofounder of WindSTAR, an NSF Industry-University Cooperative Research Center aimed at bringing together academia and industry to advance wind energy through industry-relevant research and education. Rotea joined UT Dallas in 2009 to serve as professor and inaugural head of the then newly-created mechanical engineering department. He is a Fellow of the IEEE for contributions to robust and optimal control of multivariable systems. Rotea graduated with a degree in electronic engineering from the University of Rosario. He received a master's degree in electrical engineering and his Ph.D. in control science and dynamical systems from the University of Minnesota. Dr. Rotea is a Fellow of the IEEE for contributions to robust and optimal control of multivariable systems.